\documentclass[conference]{IEEEtran}
\IEEEoverridecommandlockouts
\usepackage{cite}
\usepackage{amsmath,amssymb,amsfonts}
\usepackage{algorithmic}
\usepackage{graphicx}
\usepackage{textcomp}
\usepackage{xcolor}

\usepackage{adjustbox}

\def\BibTeX{{\rm B\kern-.05em{\sc i\kern-.025em b}\kern-.08em
    T\kern-.1667em\lower.7ex\hbox{E}\kern-.125emX}}
\begin{document}

\title{Exploring Complementarity and Explainability in CNNs for Periocular Verification Across Acquisition Distances}

\author{\IEEEauthorblockN{Fernando Alonso-Fernandez}
\IEEEauthorblockA{\textit{School of Information Technology} \\
Halmstad University, Sweden \\
feralo@hh.se}
\and
\IEEEauthorblockN{Kevin Hernandez Diaz}
\IEEEauthorblockA{\textit{School of Information Technology} \\
Halmstad University, Sweden \\
kevin.hernandez-diaz@hh.se}
\and
\IEEEauthorblockN{Jose M. Buades}
\IEEEauthorblockA{\textit{Computer Graphics and Vision and AI Group} \\
University of Balearic Islands, Spain \\
josemaria.buades@uib.es}
\and
\IEEEauthorblockN{Kiran Raja}
\IEEEauthorblockA{\textit{Faculty of Information Technology and Electrical Engineering} \\
NTNU, Norway \\
kiran.raja@ntnu.no}
\and
\IEEEauthorblockN{Josef Bigun}
\IEEEauthorblockA{\textit{School of Information Technology} \\
Halmstad University, Sweden \\
josef.bigun@hh.se}
}

\maketitle

\begin{abstract}
We study the complementarity of different CNNs for periocular verification at different distances on the UBIPr database. 
We train three architectures of increasing complexity (SqueezeNet, MobileNetv2, and ResNet50) on a large set of eye crops from VGGFace2. 
%
%Experiments are conducted on the UBIPr database using a verification protocol that includes intra- and inter-distance scenarios. 
%
We analyse performance with cosine and $\chi^2$ metrics, compare different network initialisations, and apply score-level fusion via logistic regression. 
In addition, we use LIME heatmaps and Jensen–Shannon divergence to compare attention patterns of the CNNs.
While ResNet50 consistently performs best individually, the fusion provides substantial gains, especially when combining all three networks. 
Heatmaps show that networks usually focus on distinct regions of a given image, which explains their complementarity. 
Our method significantly outperforms previous works on UBIPr, achieving a new state-of-the-art.
\end{abstract}


%\vspace{-2mm}
\section{Introduction}
%\vspace{-2mm}
The periocular region (area around the eye) is a robust biometric trait, especially under unconstrained or degraded conditions where full-face or iris capture may not be viable  \cite{Alonso24computers_periSOA}.
Compared to face or iris, it offers a balanced trade-off between accuracy, acquisition ease, and robustness to occlusion and resolution changes.
Partial face visibility is also common in contexts such as social media \cite{Hedman22_pr_selfie_beauty_filters}, masks, work gear, cultural coverings, etc. \cite{sharma23cviu_periocular_masks_survey}.
Although convolutional neural networks (CNNs) dominate feature learning in biometrics \cite{[Sundararajan18-DLbiometrics]}, their use in periocular recognition remains limited \cite{sharma23cviu_periocular_masks_survey,zanlorensi22_AIR_ocular_db_competitions_survey,zeng21_iet_FR_occlusion_survey}, partly due to the lack of large dedicated datasets \cite{zanlorensi22_AIR_ocular_db_competitions_survey}.

Several studies used off-the-shelf deep features for periocular recognition (including fusion) by leveraging networks pre-trained on ImageNet \cite{[Hernandez18],[Hernandez19],[Alonso22inffus],Hernandez23access_oneshot,Alonso24wifs_cnn_vit_ots} or face datasets \cite{[Hernandez18],tapia22_access_selfie_periocular_SR,[Alonso22inffus]}.
However, they are not specifically trained for periocular.
Other works \cite{[Zhao17],Zhao_Kumar18_tifs_periocular_attention_critical_regions,zanlorensi2022_SR_UFPR_db,Kolf22ijcb_light_ocular_lowbit_quantization,Talreja22_wacv_periocular_biometrics_leveraging_softbio,Rattani23ACCESS_OcularCNNPruningBenchmark,Kolf23_ivc_syper_ocular_db,Kolf24_eaai_MixQuantBio_face_ocular,Coelho24_lacci_PeriocularEfficientNet,Carreira24_SIBGRAPI_deep_ocular_surveillance} trained CNNs on small/mid periocular sets like UFPR (33k images) \cite{zanlorensi2022_SR_UFPR_db}, VISOB 2.0 (158k) \cite{Nguyen21_icip_visob2}, UBIPr (3.3k) \cite{[Padole12]}, or synthetic data \cite{Kolf23_ivc_syper_ocular_db}.
This contrasts with face recognition, which benefits from massive sets (e.g., WebFace260M \cite{Zhu23_pami_WebFace260M}).
Despite progress, most studies rely on a single architecture and do not examine network complementarity.
The impact of acquisition distance is also underexplored, as is attention analysis across networks, especially in multi-network setups.

Accordingly, we analyse three CNNs of varying complexity (SqueezeNet, MobileNetv2, ResNet50) for periocular recognition.
To overcome scarcity of large training sets, we use $>$1.9M ocular crops from VGGFace2 \cite{[Cao18vggface2]}, and evaluate on UBIPr \cite{[Padole12]} using intra- and inter-distance protocols.
Our CNN comparison across distances shows that ResNet50 leads individually, whereas score-level fusion via logistic regression leads to consistent improvements, especially when combining all networks.
The use of LIME heatmaps \cite{Alonso23wifs_lime_biometrics} and Jensen–Shannon divergence to visualise and quantify attention patterns reveals a clear complementarity among networks.
%
%
%Our contributions are threefold:
%
%$i)$ a CNN comparison across distances, showing that ResNet50 leads individually, with fusion offering further gains;
%
%$ii)$ score-level fusion via logistic regression, yielding consistent improvements, especially when combining all three networks; and
%
%$iii)$ the use of LIME heatmaps \cite{Alonso23wifs_lime_biometrics} and Jensen–Shannon divergence to visualise and quantify attention patterns, revealing complementarity among networks.
%
Our results show the importance of architectural diversity in enhancing performance and set a new state-of-the-art on the UBIPr dataset.

%\section{Materials and Methods}
%\vspace{-2mm}
\section{Recognition Networks}
%\vspace{-2mm}
We evaluate three CNNs of increasing complexity: SqueezeNet \cite{[Iandola16SqueezeNet]} (light, 18 layers, 1.24M params), MobileNetv2 \cite{[Sandler18mobilenetv2]} (medium, 53, 3.5M), and ResNet50 \cite{[He16]} (large, 50, 25.6M).
ResNet50 uses residual blocks that improve gradient flow and support deeper architectures. Each block reduces dimensionality with 1$\times$1 convolutions, applies 3$\times$3 filters in the reduced space, and restores the original size.
MobileNetv2 uses inverted residuals and depth-wise separable convolutions to reduce parameters and inference time. Shortcut connections link thinner layers instead, with intermediate representations in a higher-dimensional space.
%
%To reduce the number of parameters and the inference time, MobileNets employ inverted residuals, in which shortcut connections are between thinner layers instead (hence the name 'inverted'). 
%%The input and output lie in a reduced-dimensional space, whereas the intermediate representation lies in a higher-dimensional space. 
%
%Also, 3$\times$3 filters are applied via depth-wise separable convolutions, which have fewer parameters and are faster than regular filters.
%
%Thanks to the mentioned techniques, MobileNets contain the same number of convolutional layers ($\sim$50 in our case) with significantly fewer parameters.
%
SqueezeNet is a compact, non-residual network that first 'squeezes' dimensionality with 1$\times$1 filters, then 'expands' them using 1$\times$1 and 3$\times$3 convolutions in a lower-dimensional space.
This selection enables comparison of networks of varying complexity.
We adapt the ImageNet-pretrained models in MATLAB R2024b by changing the first convolution stride from 2 to 1, allowing 113$\times$113 inputs. %without altering the networks further. 
Images are normalised by subtracting 127.5 and dividing by 128.
For SqueezeNet, we follow \cite{[Alonso20SqueezeFacePoseNet]}, adding batch norm between convolutions and ReLUs, missing in the original implementation.

\begin{figure*}[t]
\centering
\includegraphics[width=0.85\textwidth]{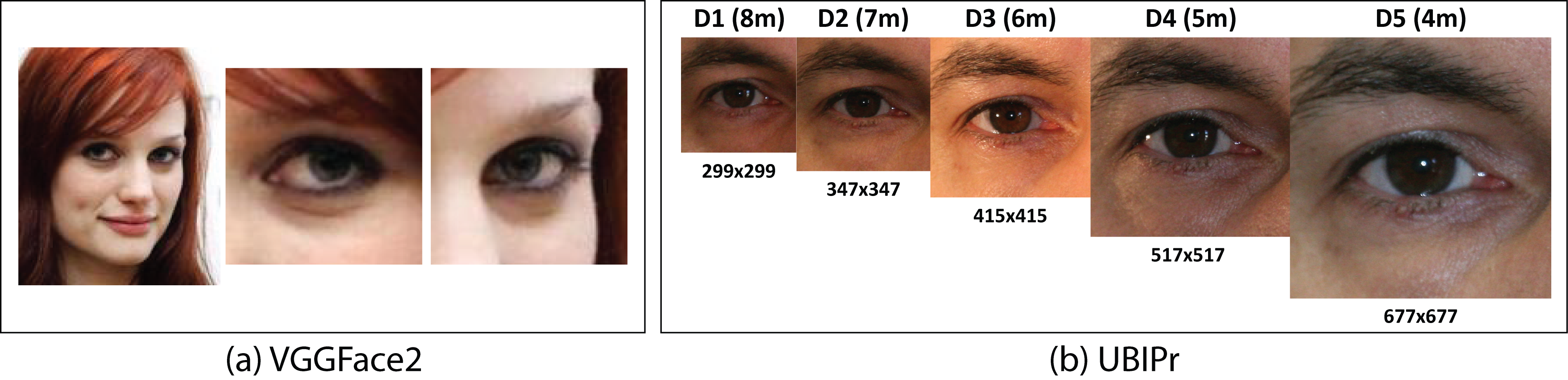}
\caption{Example images from the databases employed. The relative scale differences among normalised UBIPr images are shown, as well as their resulting size.} \label{fig:databases}
\end{figure*}

%\vspace{-2mm}
\section{Datasets}
%\vspace{-2mm}
We train using VGGFace2 (3.31M images, 9,131 identities) \cite{[Cao18vggface2]} (Figure~\ref{fig:databases}), which contains significant variation in pose, age, lighting, and background. 
Using dataset annotations, we crop ocular regions from the 8,631 training identities (3.14M images). 
%
%Ocular regions are cropped using the dataset annotations.
%
%The training set includes 8,631 classes (3.14M images).
%
Images are aligned (eye centres horizontal), scaled to 113 pixels inter-eye distance, and cropped into two 113$\times$113 patches centred on each eye.
We ensure both eyes are visible by requiring the horizontal eye-centre distance and nose vertical to be within 40\% of the inter-eye distance, and discard faces below 50 px inter-eye to avoid strong upsampling.
%
%A loose frontality check ensures that both eyes are visible by requiring the horizontal distance between eye centres and the nose vertical to be under 40\% of the inter-eye distance.
%
%Faces with inter-eye distances below 50 px are discarded to avoid excessive upsampling.
%
Left eye crops are flipped for orientation consistency, and both eyes are treated as the same identity, resulting in 953,786 valid faces and 1,907,572 ocular crops (221 per identity on average).

Testing uses the UBIPr periocular database \cite{[Padole12]}, with images from 4–8 m captured by a CANON EOS 5D.
%
%and resolutions ranging from 501$\times$401 (8m) to 1001$\times$801 pixels (4m).
%
We select 1,718 frontal images (86 subjects with two sessions, one image per eye/session/distance, totalling 86$\times$2$\times$2=344 images per distance).
%
%(except in D5, which has data from one user missing, thus having 342
%images).
%
Images are manually annotated for inner/outer eye boundaries, resized to match the average sclera radius $R_s$ of their distance group, and aligned by cropping a $7.6 R_s \times 7.6 R_s$ square around the sclera centre (Figure~\ref{fig:databases}).
Likewise, left eye crops are flipped, and both eyes represent the same identity.
Sclera boundary is used for normalisation due to its stability across dilation and better contrast than the pupil boundary. Images are finally resized to match CNN input.
Unlike methods that mask the iris \cite{[Park11]}, we retain the full periocular region to simulate realistic conditions where iris segmentation is unreliable (e.g., low resolution, blur, or low pigmentation in visible images) \cite{Alonso24computers_periSOA}.

In some cases, networks are pretrained for face recognition.
Following \cite{[Alonso20SqueezeFacePoseNet]}, ImageNet-initialised models are first trained on RetinaFace-cleaned MS1M \cite{[Guo16_MSCeleb1M]} (5.1M images, 93.4K classes), then fine-tuned on VGGFace2 (3.14M images).
As before, left eye crops are flipped, and both eyes treated as the same subject.
This two-step training leverages MS1M large volume and VGGFace2 greater intra-class diversity, having demonstrated superior performance \cite{[Cao18vggface2],[Alonso20SqueezeFacePoseNet]}.

%\begin{table}[t]
%\centering
%\caption{Verification scores with UBIPr.}
%
%\label{tab:scores}
%
%%\begin{adjustbox}{max width=0.95\textwidth}
%
%%\scalebox{0.7}{
%
%%\small
%%\footnotesize
%%\scriptsize
%%\begin{center}
%\begin{tabular}{|c|c|c|c|c|c|c|c|}
%
%\multicolumn{8}{c}{} \\ \cline{1-2} \cline{4-5} \cline{7-8}
%
%\multicolumn{2}{|c|}{\textbf{Intra-Distance}}  & \multicolumn{1}{c}{} & \multicolumn{2}{|c|}{\textbf{Inter-Distance}} & \multicolumn{1}{c}{} & \multicolumn{2}{|c|}{\textbf{Total}} \\   \cline{1-2} \cline{4-5} \cline{7-8}
%
%genuine & impostor & \multicolumn{1}{c|}{} & genuine & impostor & \multicolumn{1}{c|}{} & genuine & impostor \\  \cline{1-2} \cline{4-5} \cline{7-8}
%
%344  & 29240 & \multicolumn{1}{c|}{} & 688 &  29240  &  & 8600 & 483600 \\  \cline{1-2} \cline{4-5} \cline{7-8}
%
%%\multicolumn{8}{c}{} \\
%
%\end{tabular}
%
%%\end{center}
%
%%}
%
%%\end{adjustbox}
%
%
%\end{table}
%%\normalsize

%\vspace{-2mm}
\section{Training and Recognition Protocols}
%\vspace{-2mm}
The networks are trained for ocular identification using cross-entropy loss on VGG2 crops.
We use SGDM (batch=128, learning rate=0.01, 0.005, 0.001, 0.0001, reduced when validation plateaus).
Validation uses 2\% of training images per user.
Model weights are initialised from either ImageNet or pretrained face recognition models.
Verification is done on the UBIPr dataset for intra- and inter(cross)-distance scenarios.
Identity templates are extracted from the layer before classification (Global Average Pooling).
%
%Left and right eye images are flipped to a consistent orientation and treated as the same identity.
%
Images at distance $Di$ are compared against those at $Dj$, with $i,j \in
\left\{ {1,2,3,4,5} \right\}$. 
Intra-distance ($i=j$) genuine scores are obtained by comparing eyes of session 1 vs eyes of session 2 of the same subject (4 comparisons/user), giving 86$\times$4=344 scores per distance.
Cross-distance ($i$$\ne$$j$) scores compare eyes of session 1 at $D_i$ vs eyes of both sessions at $D_j$ (8 comparisons/user), giving 86$\times$8=688 scores per inter-distance pair.
Impostor scores are obtained by comparing eyes from session 1 of a user to eyes from session 2 of all other users, giving 86$\times$85$\times$4=29240 impostor scores per distance combination.
This results in 8600 genuine (5 intra + 10 cross) and 438600 impostor scores across 15 distance combinations.
%
%Table~\ref{tab:scores} summarises the number of score comparisons. The right-hand side of the table also presents the total number of scores for intra-distance (five combinations) and inter-distance (ten combinations) cases.
%
As comparison metrics, we use cosine similarity, commonly used in CNN-based verification, and $\chi^2$ distance, which has also demonstrated strong performance in related works \cite{[Hernandez18]}.

We also apply score-level fusion via linear logistic regression to combine multiple networks.
Given $N$ networks producing scores ($s_{1j}, s_{2j}, ... s_{Nj}$) for trial $j$, the fused score is %of these scores 
$f_j = a_0 + a_1 \cdot
s_{1j} + a_2 \cdot s_{2j} + ... + a_N \cdot s_{Nj}$,
with weights $a_{0}, a_{1}, ... a_{N}$ trained via logistic regression
\cite{[pigeon00],[brummer07fusion]}. 
This approach outperforms simple fusion rules (mean, sum) \cite{[Alonso08]} and common classifiers in multibiometrics like SVM or Random Forest \cite{[Alonso22inffus]}.
It achieved top results in ocular benchmarks through expert fusion \cite{[sequeira16crosseyed]} and recent works \cite{[Alonso22inffus],Alonso24wifs_cnn_vit_ots} involving both traditional and off-the-shelf CNNs.
Though it is a weighted sum, the coefficients are optimised with a discriminative learning rule \cite{[Bigun97]}.

%\begin{table}[t]
%\centering
%\begin{tabular}{|c|cc|cc|cc|}
%
%\multicolumn{1}{c}{} & \multicolumn{6}{c}{\textbf{Network / Initialization}} \\ \cline{2-7}
%
%\multicolumn{1}{c|}{} & \multicolumn{2}{c|}{\textbf{SQ}} & \multicolumn{2}{c|}{\textbf{MB2}} & \multicolumn{2}{c|}{\textbf{R50}} \\ \cline{2-7}
% 
%\multicolumn{1}{c|}{\textbf{Metric}} & \textbf{ImageNet} & \textbf{Face} & \textbf{ImageNet} & \textbf{Face} & \textbf{ImageNet} & \textbf{Face} \\  \hline
% 
%cosine & 5.44 & 5.97 & 2.12 & 2.24 & 1.73 & 1.95 \\  \hline
%
%$\chi^2$ & \textbf{4.93} & 5.45 & \textbf{2.10} & 2.15 & \textbf{1.66} & 1.93 \\  \hline
%
%\end{tabular}
%\caption{Ocular verification (EER \%) on UBIPr for different network initialisations (ImageNet and face recognition weights) and comparison metrics (cosine similarity and $\chi^2$ distance). The best result of each network is marked in bold. Notice the superiority of ImageNet initialisation and $\chi^2$ distance.}
%\label{tab:eer-individual-nets}
%\end{table}

\begin{table*}[t]
\centering

\begin{adjustbox}{width=0.9\textwidth}

\begin{tabular}{|c|cc|cc|cc|cc|}

\multicolumn{1}{c}{} & \multicolumn{8}{c}{\textbf{Initialization / Metric}} \\ \cline{2-9}
 
\multicolumn{1}{c|}{} & \multicolumn{4}{c|}{\textbf{ImageNet}} & \multicolumn{4}{c|}{\textbf{Face}} \\ \cline{2-9}
 
\multicolumn{1}{c|}{\textbf{network}} & \multicolumn{2}{c|}{cosine} & \multicolumn{2}{c|}{$\chi^2$} & \multicolumn{2}{c|}{cosine} & \multicolumn{2}{c|}{$\chi^2$} \\ \hline \hline

SQ & 5.44 & - & 4.93 & - & 5.97 & - & 5.45 & - \\ 
 
MB2 & 2.12 & - & 2.10 & - & 2.24 & - & 2.15 & - \\ 

R50 & \textbf{1.73} & - & \textbf{1.66} & - & \textbf{1.95} & - & \textbf{1.93} & - \\ \hline \hline

SQ+MB2 & 2.05 & (-3.07\%) & 2.04 & (-2.80\%) & 2.13 & (-4.80\%) & 2.05 & (-4.35\%) \\ 

SQ+R50 & \textbf{1.61} & (-6.96\%) & \textbf{1.51} & (-9.37\%) & 1.81 & (-7.16\%) & 1.81 & (-6.56\%) \\ 

MB2+R50 & \textbf{1.61} & (-7.27\%) & 1.59 & (-4.67\%) & \textbf{1.77} & (-8.96\%) & \textbf{1.74} & (-10.17\%) \\ \hline

ALL & \textbf{1.33} & (-23.40\%) & \textbf{1.31} & (-21.14\%) & \textbf{1.49} & (-23.20\%) & \textbf{1.50} & (-22.22\%) \\ \hline

\end{tabular}

\end{adjustbox}

\begin{adjustbox}{width=0.9\textwidth}

\begin{tabular}{|cllc|}

\multicolumn{4}{c}{} \\ 

\multicolumn{4}{c}{\textbf{Results of other works of the literature}} \\ \hline
 
\textbf{Work} & \textbf{Feature type} & \textbf{Features} & \textbf{EER} \\ \hline

\cite{[Padole12]} & traditional & SIFT+LBP+HOG &   ~16\% \\  \hline

\cite{[Alonso16a]} & traditional & SIFT+LBP+HOG &  8.4\% \\  

  & traditional & SIFT+LBP+HOG+SAFE &  7.9\% \\  \hline

\cite{[Hernandez18]} & traditional & SIFT+LBP+HOG &  9.1\% \\ \cline{2-4}

& deep &  CNN (ResNet101)   & 5.6\% \\ \cline{2-4}

& trad. + deep &  SIFT+LBP+HOG + CNN (ResNet101) &   5.1\% \\ \hline

\cite{Zhao_Kumar18_tifs_periocular_attention_critical_regions} & deep & AttNet + FCN-Peri &  2.26\% \\ \hline

\cite{Alonso24wifs_cnn_vit_ots} & trad. & SIFT+LBP+HOG &  10.58\% \\ \cline{2-4}

& deep &  CNN off-the-shelf (ResNet50)   & 8.53\% \\ \cline{2-4}

& deep &  ViT off-the-shelf (tiny)   & 11.48\% \\ \cline{2-4}

& deep &  CNN ots (ResNet50) + ViT ots (base) &   7.72\% \\  \cline{2-4}

& trad. + deep &  SIFT+LBP+HOG + CNN (ResNet50) + ViT (base) &   6.32\% \\ \hline

This work & deep & SqueezeNet + MobileNetv2 + ResNet50 &  1.31\% \\  \hline

\end{tabular}

\end{adjustbox}

\caption{Ocular verification results (EER \%) on UBIPr for different network initialisations (ImageNet and face recognition weights) and comparison metrics (cosine similarity and $\chi^2$ distance). The relative EER variation in fusion experiments with respect to the best individual network is given in brackets.  The table also shows results from previous works on the same database.}
\label{tab:eer-nets-indiv-fusion-and-other-papers}

\end{table*}

%\vspace{-2mm}
\section{Results}
%
%\subsection{Individual Networks}
%
%\vspace{-2mm}
\noindent \textbf{5.1 Individual Networks} \\
We begin by presenting (Table~\ref{tab:eer-nets-indiv-fusion-and-other-papers}, top) ocular verification results on the UBIPr database for the three networks, jointly considering intra- and inter-distance scores. 
The $\chi^2$ distance consistently provides better performance than cosine similarity, confirming earlier findings \cite{[Hernandez18]}.
While the improvement is marginal in some cases, it exceeds 0.5\% for SqueezeNet, the weakest network.
Regarding initialisation, the best case is always ImageNet. 
Although one might expect that starting from face-pretrained networks would be advantageous, since the networks are familiar with eye regions, our results suggest otherwise.
Face models may be overly specialised to full-face features, whereas ImageNet models begin with more primitive, generic features, which can better adapt to ocular data.
This supports the established view of ImageNet as a robust, versatile foundation for downstream tasks \cite{[Razavian14]}, including biometrics \cite{[Alonso22inffus],Alonso24wifs_cnn_vit_ots,[Hernandez18],[Hernandez19],Hernandez23access_oneshot,[Nguyen18]}.
In absolute performance, residual networks (MobileNetv2, ResNet50) outperform SqueezeNet, with ResNet50, the largest one, achieving the best EER at 1.66\%.

%\begin{figure}[htb]
%\centering
%    \includegraphics[width=0.96\textwidth]{C4_compute_EER_values_BOTH.eps}
%    \caption{Ocular verification results (EER \%) on UBIPr for scale variation experiments (ImageNet initialization, $\chi^2$ distance).}
%    \label{fig:eer-individual-nets-per-distance}
%\end{figure}

\begin{figure*}[t]
\centering
   \includegraphics[width=0.9\textwidth]{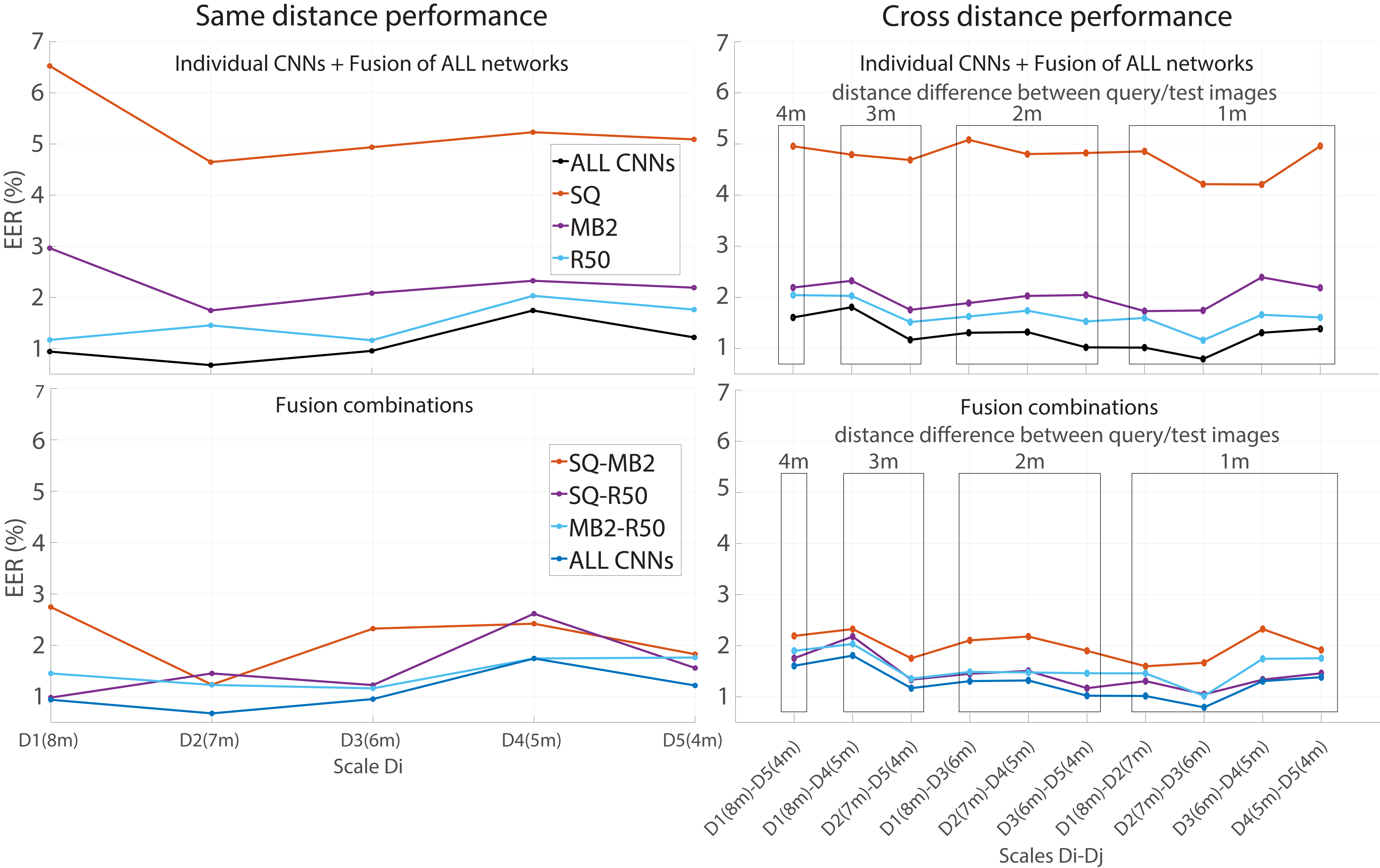}
    \caption{Ocular verification results (EER \%) on UBIPr for scale variation experiments (ImageNet initialization, $\chi^2$ distance). The figure shows the performance of the individual networks (top) and of the different fusion combinations (bottom). The top plot also shows the fusion of all networks (best fusion case) for comparison with the individual networks.}
    \label{fig:eer-individual-and-fusion-nets-per-distance}
\end{figure*}

%\begin{figure}[htb]
%\centering
%    \includegraphics[width=0.96\textwidth]{C6_compute_EER_fusion_values_BOTH_D_options_per_cnn_pairs.eps}
%    \caption{Ocular fusion verification results (EER \%) on UBIPr for scale variation experiments (ImageNet initialization, $\chi^2$ distance). The figure shows the fusion combinations separately (per row), together with the individual networks involved in the fusion.}
%    \label{fig:eer-fusion-nets-per-distance-per-cnn-pair}
%\end{figure}

%\vspace{-2mm}
%\subsection{Network Combination}
%\vspace{-2mm}
\noindent \textbf{5.2 Network Combination} \\
We then conduct (also Table~\ref{tab:eer-nets-indiv-fusion-and-other-papers}, top) fusion of the networks. % (Table~\ref{tab:eer-fusion}). %, evaluating all possible fusion combinations. 
Consistent with previous observations, both $\chi^2$  distance and ImageNet initialization remain the most effective choices.
Notably, combining two CNNs improves performance, with MobileNetv2 + ResNet50 generally being the best case. This indicates strong complementarity between these architectures, as also seen in face recognition \cite{Alonso23wifs_lime_biometrics}.
Despite both being residual networks, their differing residual layer structures likely promote diverse learned features, thus complementarity.
In contrast, fusions involving the simpler, less accurate SqueezeNet lead to smaller gains.
Only SqueezeNet with the powerful ResNet50 occasionally approaches the performance of MobileNetv2 + ResNet50.
Interestingly, fusing all three networks achieves the best performance, with improvements exceeding 20\%, compared to the 7–10\% gain when fusing just two models.
These findings highlight the advantage of leveraging architectural diversity rather than relying solely on individual model strength. 
Even lower-performing models like SqueezeNet can provide complementary information that enhances the system by compensating for the limitations of stronger models \cite{[Fierrez18],[Singh19inffus]}.

%\subsection{Comparison with Previous Works}
\noindent \textbf{5.3 Comparison with Previous Works} \\
We also report as reference in Table~\ref{tab:eer-nets-indiv-fusion-and-other-papers} (bottom) previous works on UBIPr.
Direct comparisons should be made with caution, as differences in experimental protocols occur, evidenced by the number of images used or scores reported in these works. 
A key difference in our study, shared only with \cite{Alonso24wifs_cnn_vit_ots}, is the alignment of eye crops by flipping to a common orientation and the same identity. 
This increases the number of genuine comparisons while making impostor comparisons more challenging by removing anatomical asymmetry bias.
Despite this more challenging evaluation, our results outperform all previous works.
The seminal UBIPr paper \cite{[Padole12]} set the initial benchmark, improved later by others with handcrafted features such as SIFT, LBP, HOG, or SAFE. More recent works employ deep embeddings from off-the-shelf CNNs and ViTs pretrained on ImageNet as generic feature extractors. 
In contrast, we fine-tune several networks on $>$1.9M eye crops from the large VGGFace2 dataset, achieving state-of-the-art performance on UBIPr.

\begin{figure*}[t]
\centering
    \includegraphics[width=0.9\textwidth]{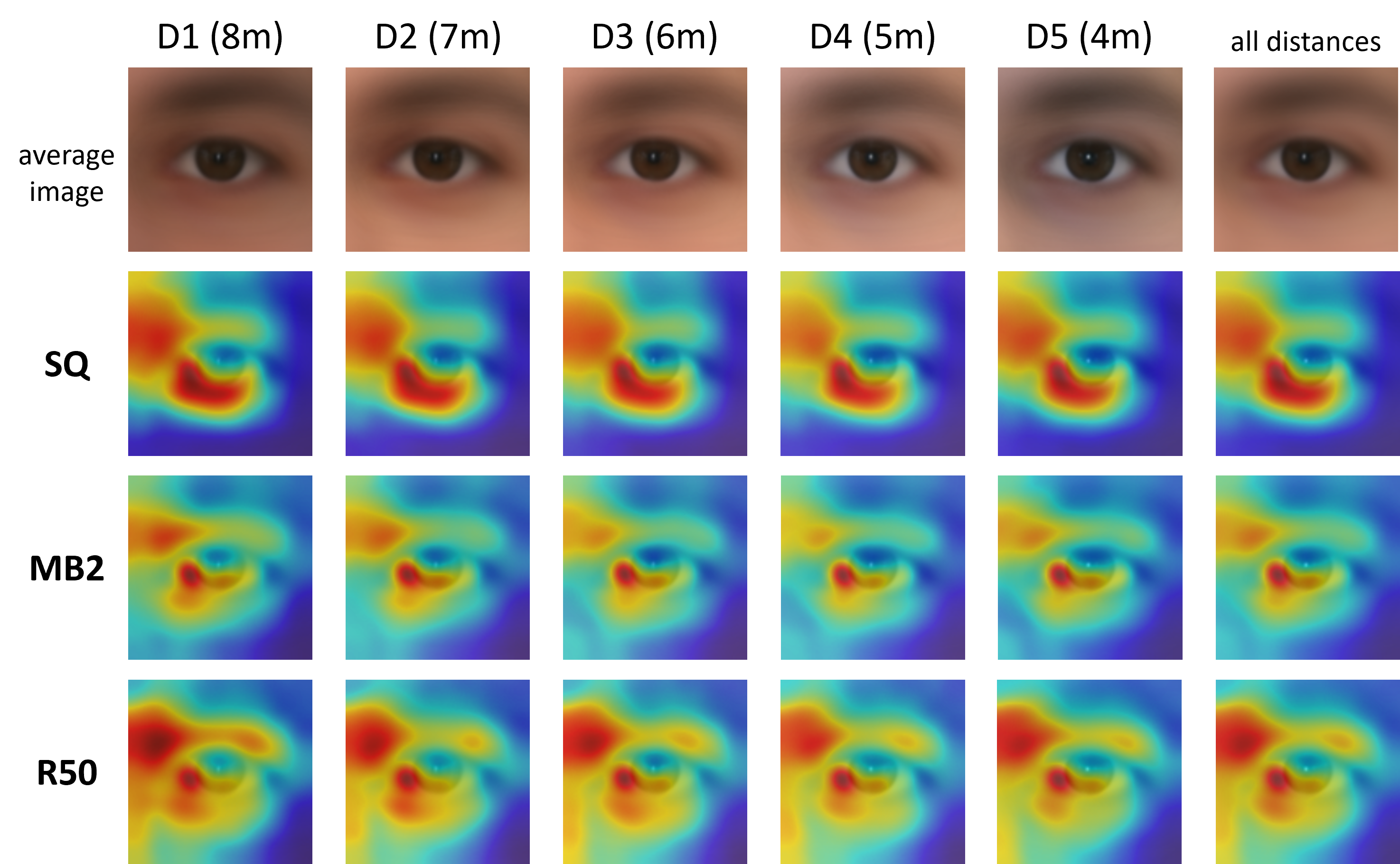}
    \caption{Average LIME heatmaps on UBIPr per distance (columns) and CNN (rows). %The top row shows the average images of the database. %The JS divergence at the bottom quantifies the similarity between the average heatmaps of all distances across CNN pairs (the lower, the more similar).
    }
    \label{fig:lime-average-maps}
\end{figure*}

%\vspace{-2mm}
%\subsection{Distance Variation}
%\vspace{-2mm}
\noindent \textbf{5.4 Distance Variation} \\
Based on the previous results, we adopt $\chi^2$ distance and ImageNet initialisation for the rest of the experiments.
We then analyse varying acquisition distances.
Figure~\ref{fig:eer-individual-and-fusion-nets-per-distance} (left column) shows the EER when both images are captured at the same distance (intra-distance), ordered from farthest (left in the x-axes) to closest (right).
The right column presents the inter-distance case, grouping results by the distance gap between image pairs, from 4 meters (left in the x-axes) to 1 meter (right).
%
%Similarly, Figure~\ref{fig:eer-individual-and-fusion-nets-per-distance} shows the corresponding EERs for different fusion combinations (top), and for the fusion of all networks compared to the individual models (bottom), as this combination provided the best overall performance (Table~\ref{tab:eer-nets-indiv-fusion-and-other-papers}).
%
%
From Figure~\ref{fig:eer-individual-and-fusion-nets-per-distance} (left column), we observe that performance is generally stable across intra-distances, except at the farthest point (8m). ResNet50 consistently performs best, with EERs below 2\%, while SqueezeNet performs worst. MobileNetv2 stays below 3\% across all ranges.
In the inter-distance setting (right column), performance degrades with increasing distance differences, especially for SqueezeNet. ResNet50 remains the most robust, with EERs under 2\% even at a 4m gap.
Fusion results confirm that combining all CNNs consistently gives the best accuracy, achieving EER$<$1.5\%, and in some cases $<$1\%, across most distance scenarios.
Fusing any two networks also improves performance, though less effectively than using all three.

\begin{figure*}[t]
\centering
    \includegraphics[width=0.9\textwidth]{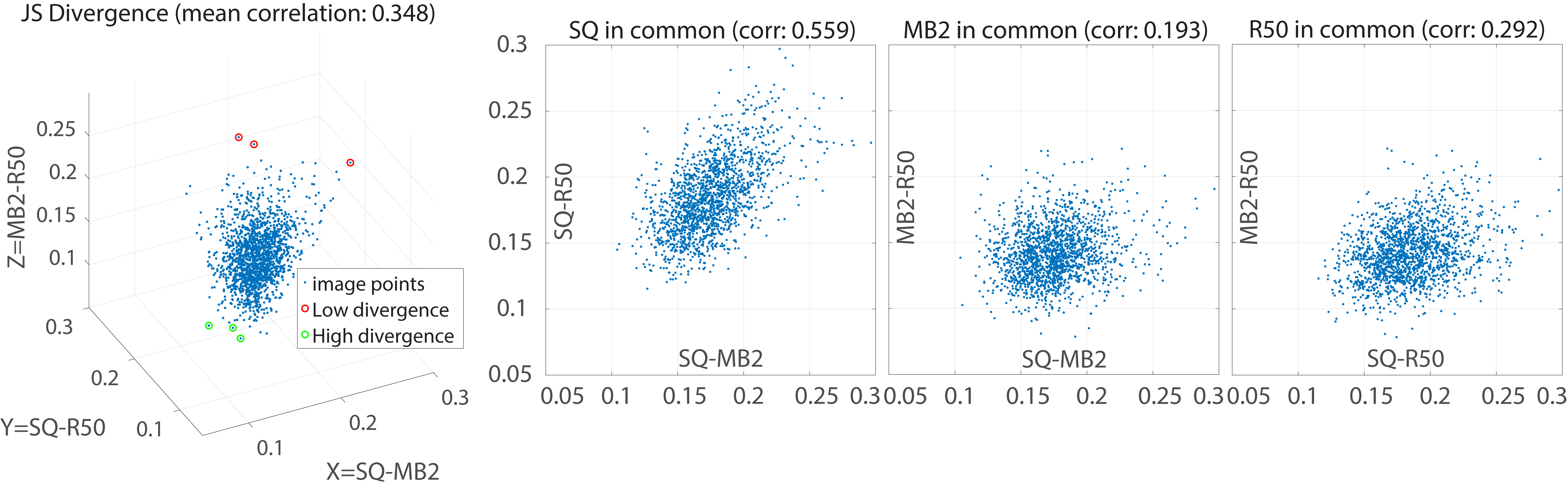}
    \caption{Jensen–Shannon divergence between the heatmaps generated by the networks. The 3D scatter plot on the left represents divergence values across images for each pair of CNNs. The three plots on the right show the 2D projections onto each pair of axes. Correlation values are also given. The clouds are computed with images of the entire database (all distances).}
    \label{fig:lime-JD-clouds}
\end{figure*}

%\vspace{-2mm}
%\subsection{Explainability Analysis}
%\vspace{-2mm}
\noindent \textbf{5.5 Explainability Analysis} \\
To further explore complementarity between networks, we analyse LIME heatmaps \cite{Alonso23wifs_lime_biometrics}, which highlight the most relevant pixels for each model.
To quantify similarity between heatmaps, we use the Jensen–Shannon divergence (JSD), a symmetric, smoothed version of the Kullback–Leibler (KL) divergence commonly used to compare probability distributions \( P \) and \( Q \). It is defined as
\(
\mathrm{JSD}(P \,\|\, Q) = 0.5 \cdot \mathrm{KL}(P \,\|\, M) + 0.5 \cdot \mathrm{KL}(Q \,\|\, M),
\)
where \( M = 0.5 \cdot (P + Q) \) is the average distribution, and % \( \mathrm{KL}(P \,\|\, Q) \) denotes the Kullback--Leibler divergence 
\(
\mathrm{KL}(P \,\|\, Q) = \sum_i P(i) \log (P(i)/Q(i))
\)
is the standard KL divergence.
Heatmaps are normalized into probability distributions by dividing each pixel by the total sum.
JSD ranges from 0 (identical) to $\log(2) \approx 0.6931$ (maximally different distributions).
Figure~\ref{fig:lime-average-maps} presents the average LIME heatmaps of each network at different distances.
%
%The top row shows the average image per distance, while the bottom part presents the Jensen–Shannon divergence (JSD) values computed between the average heatmaps of CNN pairs, quantifying the similarity between their attention distributions.
%
Recall that left eyes are horizontally flipped for orientation consistency (nose on the left).
Overall, MobileNetv2 has more localised and compact activations, particularly under the lower eyelid, while ResNet50 and SqueezeNet show broader patterns.
All models highlight regions like the upper eyelid, sclera, and tear duct, especially ResNet50 and SqueezeNet.
The cheek and right periocular part receive minimal attention, and interestingly, the pupil/iris is also less attended, suggesting reliance on periocular context rather than iris texture. 
Across distances, the heatmaps remain relatively stable within each network, in line with the consistent performance seen in Figure~\ref{fig:eer-individual-and-fusion-nets-per-distance} (left).

\begin{figure*}[t]
\centering
    \includegraphics[width=0.9\textwidth]{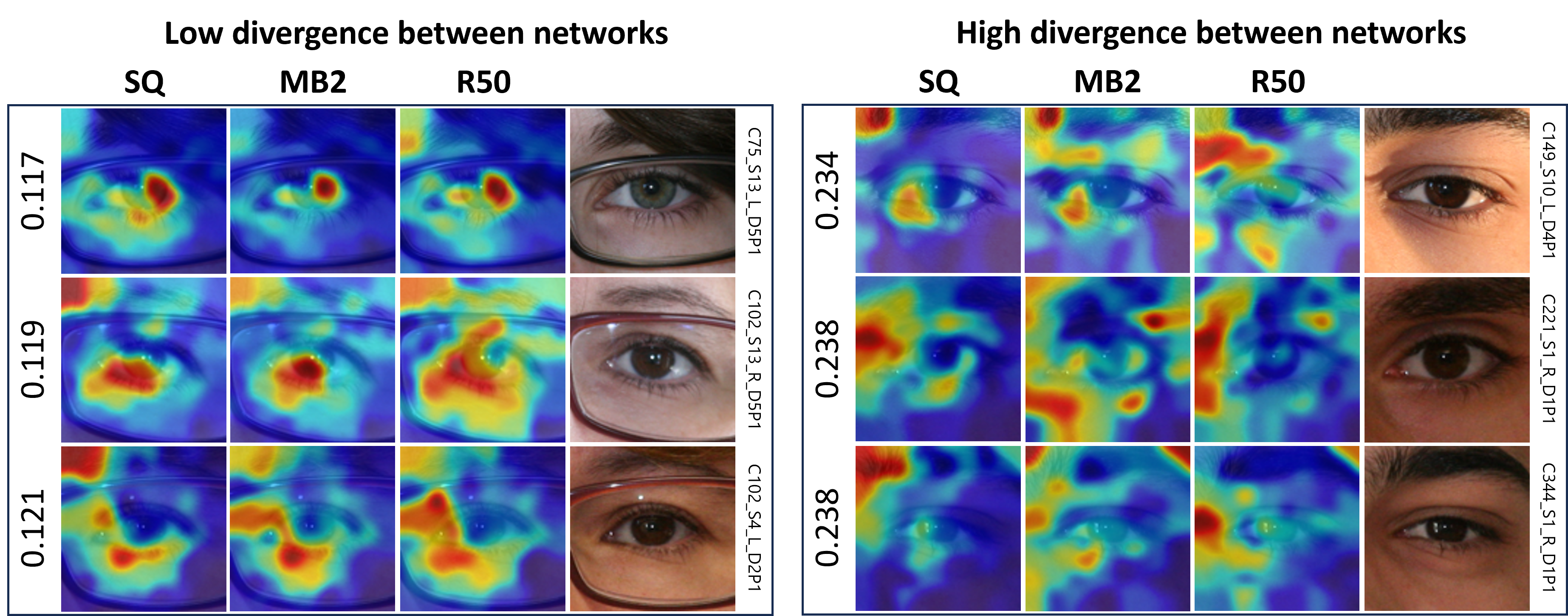}
    \caption{Individual heatmaps where the three networks diverge the least (left) and the most (right). The numeric values indicate the average JS divergence between the three possible network pairs.}
    \label{fig:lime-extreme-diverging-maps}
\end{figure*}

While average heatmaps reveal common attended regions, we also assess differences per image. 
To do so, we compute the pairwise JSD between networks for each image and plot the results in a 3D scatter space (Figure~\ref{fig:lime-JD-clouds}), where each axis represents JSD for a specific network pair. 
The cloud shape suggests low correlation between divergence values, indicating that the networks often produce complementary explanations.
In particular, 2D projections onto SQ–MB2 vs. MB2–R50 and SQ–R50 vs. MB2–R50 planes are near-circular, with low Pearson correlations (0.193, 0.292). 
A more linear trend appears in SQ–MB2 vs. SQ–R50, with a moderate correlation of 0.559, suggesting that SqueezeNet tends to agree/disagree similarly with the other two CNNs.
However, the overall lack of strong correlation supports the benefit of fusing networks with distinct attention patterns.
Figure~\ref{fig:lime-extreme-diverging-maps} shows examples with the lowest and highest average JSD between network pairs (marked in Figure~\ref{fig:lime-JD-clouds}).
Interestingly, the lowest-divergence cases often involve glasses, which the networks learn to ignore, focusing on consistent periocular features such as the skin below the lower eyelid, tear duct, and sclera. 
Regarding the highest-divergence cases, they show varied attention to the sclera, tear duct, lower eyelid, or eyelashes.

%\clearpage
%
%\begin{table}[htb]
%\centering
%\begin{tabular}{|c|c|c|c|c}
%\cline{1-4}
%\begin{tabular}[c]{@{}c@{}}common\\ network:\end{tabular} & SQ & MB2 & R50 &  \\ \hline
%\begin{tabular}[c]{@{}c@{}}network\\ pairs:\end{tabular} & \begin{tabular}[c]{@{}c@{}}SQ-MB2 vs.\\ SQ-R50\end{tabular} & \begin{tabular}[c]{@{}c@{}}SQ-MB2 vs.\\ MB2-R50\end{tabular} & \begin{tabular}[c]{@{}c@{}}SQ-R50 vs.\\ MB2-R50\end{tabular} & \multicolumn{1}{c|}{\begin{tabular}[c]{@{}c@{}}average\\ correlation\end{tabular}} \\ \hline
%\textbf{D1 (8m)} & 0.584 & 0.163 & 0.29 & \multicolumn{1}{c|}{0.346} \\ \hline
%\textbf{D2 (7m)} & 0.54 & 0.238 & 0.244 & \multicolumn{1}{c|}{0.341} \\ \hline
%\textbf{D3 (6m)} & 0.586 & 0.242 & 0.307 & \multicolumn{1}{c|}{0.378} \\ \hline
%\textbf{D4 (5m)} & 0.579 & 0.227 & 0.299 & \multicolumn{1}{c|}{0.368} \\ \hline
%\textbf{D5 (4m)} & 0.518 & 0.101 & 0.327 & \multicolumn{1}{c|}{0.315} \\ \hline
%\textbf{all} & 0.559 & 0.193 & 0.292 & \multicolumn{1}{c|}{0.348} \\ \hline
%\end{tabular}
%\caption{Correlation of the heatmaps divergence between CNN pairs (per distance). The last row shows values of the entire database (all distances), matching the numbers shown in Figure~\ref{fig:lime-JD-clouds}.}
%\label{tab:lime-correlations-distances}
%\end{table}

%\vspace{-2mm}
\section{Conclusions}
%\vspace{-2mm}
This work analysed the performance and complementarity of three CNNs of different complexity and depth for periocular verification under scale variation on the UBIPR database \cite{[Padole12]}. We observed that deeper, residual networks (e.g., ResNet50) perform best individually, but the best results are obtained when combining all three models. Score-level fusion using logistic regression provided up to 23\% relative improvement over the best network.
Using LIME-based heatmaps \cite{Alonso23wifs_lime_biometrics} and Jensen–Shannon divergence, we further showed that each network focuses on different regions of the eye, suggesting that their feature representations are complementary. This explains the success of the fusion strategy and support the use of explainability tools to guide architectural decisions. Our method establishes new state-of-the-art results on the UBIPr dataset and demonstrates the value of combining diverse CNN architectures for robust periocular verification.

Despite being captured at several distances, UBIPr contains high-resolution images given by a CANON EOS 5D camera (22.3 MPx) and cooperative subjects. 
It remains to be seen how well the system generalizes to more challenging scenarios, including lower-quality sensors such as those used in surveillance environments, and non-cooperative subjects.
We would also like to test our approach in near-infrared data, for which spectrum translation techniques may be employed \cite{HernandezDiaz_20_ijcb_cross_spectral_ocular_translation} to ensure sufficient training data in this domain as well.
We are also working on integrating more discriminative loss functions, such as margin-based approaches like ArcFace.
Another avenue is the adoption of a sequential fine-tuning strategy, where networks are first trained on ocular crops from MS-Celeb-1M (MS1M) and later refined using VGGFace2. This approach can exploit the larger scale of MS1M for initial generalization and benefit from the greater intra-class variability in VGGFace2, an strategy seen very effective strategy in face recognition \cite{[Cao18vggface2],[Alonso20SqueezeFacePoseNet]}.

%\footnotesize

\noindent \textbf{Acknowledgements}. 
This work was partly done while F. A.-F. was visiting researcher at University of Balearic Islands.
F. A.-F., K. H.-D., and J. B. thank the
Swedish Research Council (VR) and the EU (HORIZON Europe project PopEye under Grant Agreement no 101168317). % for funding their research.
Funded by the European Union. Views and opinions expressed are, however, those of the author(s) only and do not necessarily reflect those of the European Union or the European Research Executive Agency. Neither the European Union nor the granting authority can be held responsible for them.
This work is part of the Project PID2022-136779OB-C32 (PLEISAR) funded by MICIU/ AEI /10.13039/501100011033/ and FEDER, EU.
%
%Author J. M. B. thanks the project EXPLAINING - "Project EXPLainable Artificial INtelligence systems for health and well-beING", under Spanish national projects funding (PID2019-104829RA-I00/AEI/10.13039/501100011033).

\bibliographystyle{IEEEtran}
\bibliography{fernando1_biosig25}

\end{document}